\pdfoutput=1

\documentclass[11pt]{article}

\usepackage[preprint]{acl}

\usepackage{times}
\usepackage{latexsym}

\usepackage[T1]{fontenc}

\usepackage[utf8]{inputenc}

\usepackage{microtype}

\usepackage{inconsolata}

\usepackage{graphicx}

\usepackage{multirow}
\usepackage{multicol}
\usepackage{booktabs}
\usepackage{tabularx}
\usepackage{makecell}
\usepackage{lscape} 
\usepackage{enumitem}
\setitemize{topsep=3pt,parsep=1pt,partopsep=1pt}
\usepackage[export]{adjustbox}
\usepackage{float}

\title{CIVET: Systematic Evaluation of Understanding in VLMs}

\author{First Author \\
  Affiliation / Address line 1 \\
  Affiliation / Address line 2 \\
  Affiliation / Address line 3 \\
  \texttt{email@domain} \\\And
  Second Author \\
  Affiliation / Address line 1 \\
  Affiliation / Address line 2 \\
  Affiliation / Address line 3 \\
  \texttt{email@domain} \\}

\makeatletter
\def\@fnsymbol#1{
    \ensuremath{
        \ifcase#1\or \dagger\or \ddagger\or \mathsection\or \mathparagraph\or \|\or **\or \dagger\dagger \or \ddagger\ddagger \else\@ctrerr\fi
    }
}
\makeatother

\author{
    Massimo Rizzoli\textsuperscript{\ $\dagger$\ }, Simone Alghisi\thanks{Equal contribution.}, Olha Khomyn, Gabriel Roccabruna, \\ 
    \textbf{Seyed Mahed Mousavi, Giuseppe Riccardi} \\
     Signals and Interactive Systems Lab, University of Trento, Italy \\
    \texttt{ \{massimo.rizzoli, s.alghisi, giuseppe.riccardi\}@unitn.it}
}

\begin{document}

\newcommand{\llava}[0]{LLaVA-NeXT}
\newcommand{\molmo}[0]{Molmo}
\newcommand{\qwen}[0]{Qwen2-VL}
\newcommand{\clip}[0]{CLIP}
\newcommand{\llavas}[0]{LLaVA-NeXT }
\newcommand{\molmos}[0]{Molmo }
\newcommand{\qwens}[0]{Qwen2-VL }
\newcommand{\clips}[0]{CLIP }
\newcommand{\logo}[0]{\includegraphics[height=2.75ex,valign=b, trim={6.5cm 12.5cm 6.5cm 12.5cm}, clip=true]{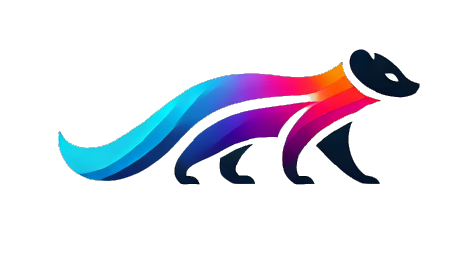}}

\maketitle
\begin{abstract}
While Vision-Language Models (VLMs) have achieved competitive performance in various tasks, their comprehension of the underlying structure and semantics of a scene remains understudied.
To investigate the understanding of VLMs, we study their capability regarding object properties and relations in a controlled and interpretable manner. 
To this scope, we introduce CIVET\footnote{We release all the materials of CIVET and encourage the community to extend this framework and its components for different evaluation and training settings: \url{https://github.com/sislab-unitn/CIVET}.}\logo, a novel and extensible framework for systemati\textbf{C} evaluat\textbf{I}on \textbf{V}ia controll\textbf{E}d s\textbf{T}imuli.
CIVET addresses the lack of standardized systematic evaluation for assessing VLMs' understanding, enabling researchers to test hypotheses with statistical rigor.
With CIVET, we evaluate five state-of-the-art VLMs on exhaustive sets of stimuli, free from annotation noise, dataset-specific biases, and uncontrolled scene complexity.
Our findings reveal that 1) current VLMs can accurately recognize only a limited set of basic object properties; 2) their performance heavily depends on the position of the object in the scene; 3) they struggle to understand basic relations among objects. Furthermore, a comparative evaluation with human annotators reveals that VLMs still fall short of achieving human-level accuracy.
\end{abstract}

\begin{figure}[t!]
    \centering
    \includegraphics[width=0.9\columnwidth]{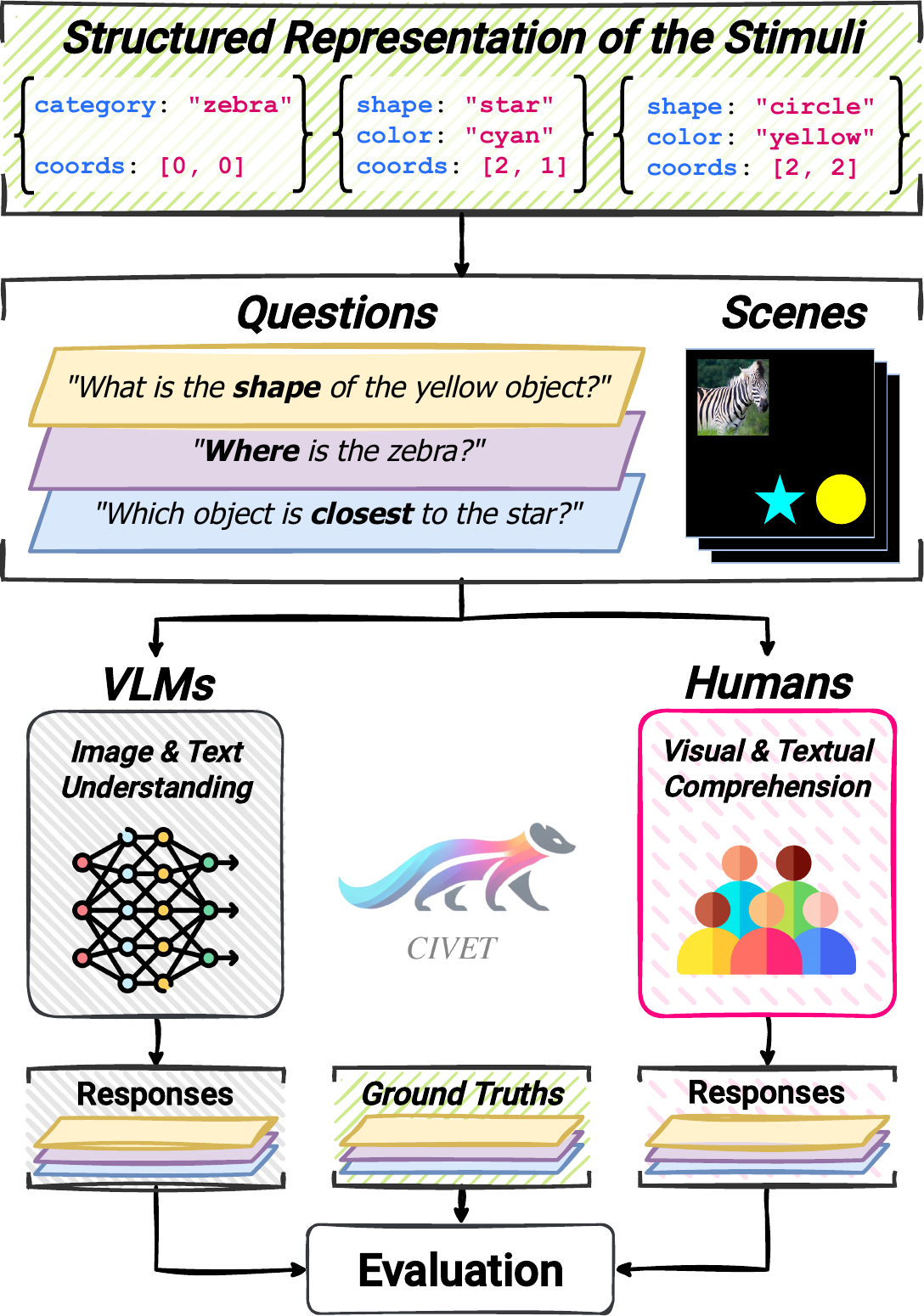}
    \caption{
        CIVET framework enables a systematic evaluation of VLMs. The framework includes customizable Stimuli, and their Structured Representation; deterministic generation of Stimuli instances (Questions and corresponding visual Scenes); and comparative assessment of VLMs and Humans understanding.
    }
    \label{fig:framework}
\end{figure}

\section{Introduction}
Recent advancements have shown that Vision-Language Models (VLMs) have achieved competitive performance on several vision-language tasks.
However, data used to train and evaluate these models is limited in size, and may suffer from annotation errors \cite{schuhmann2021laion}, label imbalance \cite{acharya2019tallyqa}, visual scene biases (e.g. objects often in the center \cite{kirillov2023segment}), and scene complexity (e.g. number and type of objects \cite{lin2014microsoft}).
This may affect evaluation results and the outcome of the learning process, hindering VLMs' generalization capabilities and providing data- or task-specific performance.
Indeed, studies on object classification with randomly augmented \cite{roth2023waffling} and spurious non-visual descriptions \cite{esfandiarpoor-etal-2024-clip} suggest that VLMs may exploit dataset biases and statistical shortcuts rather than understanding the underlying structure and semantics of the scene.

Different from previous works that focus on solving specific tasks using limited data \cite{thrush2022winoground, paiss2023teachclip-countbench, fu-etal-2023-gen-then-select, chen2024spatialvlm}, we conduct a broader and exhaustive investigation into VLMs' understanding. 
Specifically, we frame our study on three research questions: 
\textbf{RQ1: Can VLMs accurately recognize basic object properties?} Recognizing object properties is essential to distinguishing between similar elements of a scene and generalizing to previously unseen objects using known attributes. 
\textbf{RQ2: Is their performance robust to variations in object positioning?} Although a model may recognize different object properties, understanding requires consistent performance even when other variables change, such as object position.
\textbf{RQ3: Can VLMs identify basic relations among objects?} To support reasoning and understanding of visual scenes, models must go beyond recognizing individual objects and properties, and capture how objects relate to each other and interact in the world.

Answering these questions rigorously requires a systematic evaluation that includes carefully designed and controlled stimuli.
However, available evaluation frameworks based on generative models \cite{Peng_2024_CVPR} may suffer from hallucinations and lack the scalability required for systematic experimentation.
Meanwhile, earlier deterministic approaches \cite{Andreas_2016_CVPR, johnson2017clevr} — while more controllable — were not designed to ensure uniform distributions of visual scenes, precluding systematic evaluation. Additionally, since these stimuli may have been included in VLMs’ training data, their reliability as evaluation tools is further compromised.

To address these issues, we introduce CIVET \logo, a novel and extensible framework for systemati\textbf{C} evaluat\textbf{I}on \textbf{V}ia controll\textbf{E}d s\textbf{T}imuli.
Unlike previous work, CIVET allows systematic evaluation with statistical guarantees, free from external confounding factors, achieved with precise control over the content of visual and textual stimuli and deterministic generation.
Using CIVET, we systematically evaluate the performance of five state-of-the-art VLMs in recognizing properties and relations of elementary objects by generating tailored sets of stimuli, balanced in terms of position, property values, and labels.
Since these elementary objects might overestimate performance, we also evaluate the VLMs on real-world objects (i.e., objects from MS COCO \cite{lin2014microsoft}).
Finally, we conduct a study with human annotators and compare their performance to that of the VLMs. Figure \ref{fig:framework} shows an overview of the CIVET framework.

In summary, the main contributions of this paper are:
\begin{itemize}
    \item CIVET\logo, a novel and extensible framework to systematically evaluate VLMs' understanding;
    \item Exhaustive evaluation of the understanding of five state-of-the-art VLMs in recognizing object properties and relations;
    \item Comparative evaluation of VLMs and humans' performance on the same set of stimuli.
\end{itemize}

\section{Literature Review}
Several works have evaluated VLMs across a wide range of tasks, including VQA \cite{goyal2017making, yue2023mmmu, chen2024spatialvlm}, reasoning with external knowledge \cite{fu-etal-2022-time-and-place, fu-etal-2023-gen-then-select}, counting \cite{acharya2019tallyqa, paiss2023teachclip-countbench}, and understanding object relations \cite{krishna2017visualgenome, thrush2022winoground, yuksekgonul2022ARO}.
These works focus on specific tasks without assessing VLMs' understanding. 
Indeed, their evaluation is based on real-world visual scenes paired with human-annotated ground truths, which often suffer from issues such as label imbalance \cite{acharya2019tallyqa}, positional biases (e.g., relevant objects appearing centrally \cite{kirillov2023segment}), and unknown or uncontrolled scene complexity, including occlusions, distractors, and ill-posed questions.
To address some of these limitations, other frameworks have been proposed to assess the performance on visual-language tasks in a controlled setting. 
While SPEC \cite{Peng_2024_CVPR} proposed to study VLMs' understanding of objects' properties and relations by generating realistic visual scenes, it uses diffusion models, which are known to suffer from hallucination \cite{aithal2024understanding-hallucinations, kim2024structural-hallucination}.
Unlike CIVET, SPEC provides no guarantee of being free of annotation error, hindering the interpretability of results.
On the other hand, earlier deterministic approaches, such as SHAPES \cite{Andreas_2016_CVPR} and CLEVR \cite{johnson2017clevr}, eliminate annotation errors but do not ensure uniform distributions of visual scenes, precluding systematic evaluation.
Additionally, since the (pre-)training data of VLMs is often undisclosed, they may have been partially included in the pretraining.
This makes them unsuitable to answer our research questions, as good performance may not reflect VLMs' scene understanding.
Furthermore, the 3D scenes of CLEVR include additional complexity (e.g., occlusions, reflections, and shadows) that may confound our evaluation.

\section{CIVET Framework \logo}
We introduce CIVET, a framework designed to address the lack of standardized, systematic evaluation to assess VLMs' understanding.
CIVET enables systematic investigation of open research questions in the field by leveraging an exhaustive set of controlled visual scenes and natural language inputs.
We formalize the framework to make it extensible, allowing researchers to adapt it to diverse evaluation objectives.
Following the definitions given in Ontology \cite{sep-object}, an \textit{object} is defined as an instance of an \textit{entity}\footnote{Where an entity is \textit{``independent, separate, or self-contained existence''}, from the Merriam-Webster dictionary.} and is characterized by a given set of properties.
A \textit{property} is a characteristic of an object, such as its shape or color.
Moreover, the way objects stand to each other is called a \textit{relation}, such as their relative position (e.g., on top, or in front), or relative size (e.g., smaller, or larger).
We define a world as a set of objects and their relations, where objects are characterized by property-value pairs.
Each world is a structured representation of a stimulus, which can be used to generate a scene (i.e., a visual representation of the world), and a set of natural language questions about its objects, properties, and relations.
To evaluate the understanding of an aspect of the scene, we fix that aspect and marginalize over all combinations of the remaining variables. 
For example, to assess recognition of a particular shape like a star, we consider all scenes that contain a star, and marginalize over variations in color and position.
This allows us to isolate the model’s understanding of shape by averaging out the influence of other factors.

\section{Experimental Settings}
To rigorously answer our research questions, we need tailored sets of stimuli that are free from annotation error and visual biases, and are balanced in terms of position, property values, and labels.
Using CIVET, we systematically evaluate the understanding of five VLMs by generating these controlled sets of stimuli.

\subsection{Settings}
\label{sec:dataset}
We design five settings to address our research questions: \textbf{Single Object} and \textbf{Single Object w. COCO} for the recognition of object properties (RQ1) and independence to object position (RQ2); and \textbf{Relative Position}, \textbf{Relative Size}, and \textbf{Relative Distance} for the recognition of relations among objects (RQ3).
In all settings, the task requires answering closed-ended questions (Table \ref{tab:question-templates}) about a scene.
Each scene is a $9 \times 9$ grid that corresponds to the visual representation of the world (containing its set of objects).
For each question, we provide the set of possible answer options by appending to the question \textit{``Choose from [<options>]''} (where \textit{<options>} is a comma-separated list of all possible answer options to the question).
We limit the order bias by shuffling the options so that each possible order appears uniformly in the textual input.
Additionally, as models tended to respond with open answers, we condition the models prepending the instruction \textit{"Answer with as few words as possible."}.
We discuss this solution in detail in Appendix \ref{subsec:answ-length}.

\begin{table}[t]
    \centering
    \small
    \begin{tabular}{lp{4.5cm}}
        \toprule
        \textbf{Experiment} & \textbf{Question Template} \\
        \midrule
        Properties    &	\textit{What is the <property> of the object?}	\\
        \midrule
        Absolute Position &	\textit{Where is the <sheen> <color> <shape>?}	\\
        \midrule
        Relative Position &	\textit{Where is the <shape\textsubscript{1}> positioned with respect to the <shape\textsubscript{2}>?}	\\
        \midrule
        Relative Distance &	\textit{What is the closest object to the <shape>?}	\\
        \midrule
        Relative Size &	\textit{What is the size of the <shape\textsubscript{1}> with respect to the <shape\textsubscript{2}>?}	\\
        \bottomrule
    \end{tabular}
    \caption{
        Natural language question templates used in each experiment.
        We make the questions closed-ended by appending \textit{"Choose from [<options>]."}, and replacing \textit{<options>} with the corresponding answer options.
    }
    \label{tab:question-templates}
\end{table}

\paragraph{Single Object (RQ1, RQ2)}
\label{par:single-obj}
To answer our questions about the model's ability to recognize object properties (i.e., \textit{shape}, \textit{color}, \textit{sheen}) and its position w.r.t. the background (i.e., \textit{absolute position}), we consider worlds containing exactly a single object, eliminating other confounding factors.
As objects, we consider all the combinations of 4 shapes (\textit{square}, \textit{circle}, \textit{triangle}, \textit{star}), 6 colors (\textit{red}, \textit{green}, \textit{blue}, \textit{cyan}, \textit{magenta}, \textit{yellow}), and 3 values for sheen (either \textit{no sheen}, or \textit{matte} or \textit{glossy} sheen).
Then, for each object we create 81 different visual scenes by placing it in each possible cell of our $9 \times 9$ grid (for a total of 5,832 scenes).
When querying about the absolute position of the object, we divide the scene into 9 equal sections of $3 \times 3$ cells.
Then, we assign to each section (top-to-bottom, left-to-right) a unique label (i.e., \textit{top left}, \textit{top center}, \textit{top right}, \textit{center left}, \textit{center}, \textit{center right}, \textit{bottom left}, \textit{bottom center}, \textit{bottom right}) and use them as ground truth.

\begin{table}[t]
    \centering
    \small
    \begin{tabularx}{0.70\columnwidth}{lrr}
        \toprule
        \textbf{M\scriptsize{odel}} & \textbf{C\scriptsize{ategory}} & \textbf{P\scriptsize{osition}} \\
        \midrule
        \textit{Random Baseline}
        &	33	&	11\\
        \midrule
        \textit{LLaVA-NeXT 7B}
        &	91	&	37\\
        \midrule
        \textit{LLaVA-NeXT 13B}
        &	80	&	51\\
        \midrule
        \textit{Molmo-O 7B}
        &	70	&	\textbf{53}\\
        \midrule
        \textit{Qwen2-VL 7B}
        &	\textbf{97}	&	52\\
        \midrule
        \textit{CLIP}
        &	67	&	15\\
        \bottomrule
    \end{tabularx}
    \caption{Accuracy (\%) of each model when considering visual scenes containing a Since Object extracted from COCO (among \textit{zebra}, \textit{giraffe}, and \textit{elephant}) and querying about their \textit{category} and \textit{absolute position} (w.r.t. the background). Results are based on $1344 \times 1344$ images.}
    \label{tab:coco-rq1-acc}
\end{table}

\paragraph{Single Object w. COCO (RQ1, RQ2)}
Since elementary objects might overestimate VLM performance, we complement synthetic worlds with real-world objects extracted from COCO images \cite{lin2014microsoft}.
For each bounding box, we use CLIP to classify the category of the contained object.
We select the three object categories with the best performance, namely \textit{giraffe}, \textit{elephant}, \textit{zebra}.
For each category, we analyze the 10 objects with the highest similarity (dot product) to their category and manually select the best one (i.e., containing a single, non-occluded instance of the category).
Similarly to its synthetic counterpart, we answer our questions about the model's capability to recognize the object properties (i.e., \textit{category}) and its position w.r.t. the background (i.e., \textit{absolute position}) by designing a set of stimuli containing a single object. 
We consider all combinations of 3 categories (\textit{giraffe}, \textit{elephant}, \textit{zebra}) and cell placement in our $9\times9$ grid (for a total of 243 scenes).

\paragraph{Relative Position (RQ3)}
To understand whether VLMs can identify basic relations among multiple objects, we first assess their performance on \textit{relative position}, which concerns the placements of objects in the scene to one another.
Based on the results of the previous experiments, we select two \textit{yellow} objects with different shapes as discriminants (i.e., \textit{yellow} \textit{star} and \textit{yellow} \textit{triangle}).
We then construct a set of stimuli by placing the two objects in all combinations of cells (for a total of 6,480 visual scenes) and query about the object's relative position.
As ground truths, we consider 8 possible answers: 4 where the objects are on the same row or column (\textit{directly above}, \textit{directly left}, \textit{directly right}, \textit{directly below}), and 4 where the object are offset on both row and column (\textit{above left}, \textit{above right}, \textit{bottom left}, \textit{bottom right})

\paragraph{Relative Size (RQ3)}
\label{par:rel-size}
As a second type of relation, we assess the performance of VLMs in recognizing the \textit{relative size} of an object w.r.t. another.
In this setting, we consider four \textit{yellow} objects with different shape and size\footnote{The \textit{small} size is a quarter of the \textit{regular} size (half the width and height).} combinations (i.e., \textit{regular} \textit{yellow} \textit{star}, \textit{small} \textit{yellow} \textit{star}, \textit{regular} \textit{yellow} \textit{triangle},  and \textit{small} \textit{yellow} \textit{triangle}).
For each combination of two objects, we then generate one visual scene for each pair of cells (for a total of 25,920).

\paragraph{Relative Distance (RQ3)}
Finally, we assess their performance on relative distance.
We design a setting considering three \textit{yellow} objects with different shapes (i.e., \textit{yellow} \textit{star}, \textit{yellow} \textit{triangle}, and \textit{yellow} \textit{circle}).
Due to the large number of combinations, we place each object in one of 9 sections (see Single Object in Section \ref{par:single-obj}).
For each resulting configuration, we sample the cell uniformly from the section (for a total of 4,374 scenes).

\begin{table}[t]
    \centering
    \small
    \begin{tabularx}{0.95\columnwidth}{lrrrrrr}
        \toprule
        \textbf{M\scriptsize{odel}} & \textbf{S\scriptsize{hape}} & \textbf{C\scriptsize{olor}} & \textbf{S\scriptsize{heen}} & \textbf{P\scriptsize{osition}} \\
        \midrule
        \textit{Random Baseline}
        &	25	&	17	&	50	&	11\\
        \midrule
        \textit{LLaVA-NeXT  7B}  
        &	98	&	88	&	50	&	42\\
        \midrule
        \textit{LLaVA-NeXT  13B}	            
        &	97	&	76	&	\textbf{64}	&	47\\
        \midrule
        \textit{Molmo-O 7B}
        &	\textbf{100}	&	98	&	59	&	\textbf{62}\\
        \midrule
        \textit{Qwen2-VL 7B}
        &	99	&	\textbf{99}	&	60	&	61\\
        \midrule
        \textit{CLIP}
        &	95	&	95	&	49	&	14\\
        \bottomrule
    \end{tabularx}
    \caption{Accuracy (\%) of each model when considering visual scenes containing a Single Object and querying about its \textit{shape}, \textit{color}, \textit{sheen}, and \textit{absolute position} (w.r.t. the background). Results are based on $672 \times 672$ images.}
    \label{tab:rq1-acc}
\end{table}

\subsection{Models}
We select representative VLMs covering several architectures and training strategies: LLaVA-NeXT 7B \cite{liu2024llavanext}, Molmo 7B-O \cite{deitke2024molmo}, and Qwen2-VL-7B-Instruct \cite{wang2024qwen2}.
To study whether scaling the text decoder affects the performance, we also evaluate LLaVA-NeXT 13B.
As it is used as the vision encoder for \llavas and \molmo, we also consider CLIP ViT-L/14-336px \cite{radford2021clip} to understand its contribution on the performance of the VLMs.
Since \clips was trained to maximize the similarity between text and an image, we map our closed-ended question-answering task into a classification task, where the classes are the answer options.
We follow standard practice for zero-shot image classification with \clips \cite{radford2021clip}, i.e., we use \clips to encode the image and each option, and select the option with the highest similarity to the image.

In general, most VLMs include a vision encoder and a text decoder, but they differ in how they combine these components and present visual information to the text decoder.
Both \llavas and \molmos combine their pre-trained encoder and decoder using a projection layer. 
However, \llavas only trains the projection layer, while \molmos fine-tunes the whole architecture.
\qwens instead trains the vision encoder and the text decoder jointly, forcing them to learn a shared feature representation (without the need for a projection layer).
Regarding the vision encoder, \llavas and \molmos rely on the pre-trained vision encoder of \clip, which can only handle images of exactly $336\times336$ pixels.
For higher-resolution images, these models must either resize or subdivide images into smaller patches, potentially losing detail.
Differently, the vision encoder of \qwens natively handles images of different resolutions, without any resizing.
Additional details about the GPU requirements and the models are available in Appendix \ref{sec:exp-details}.

\section{Evaluation}
We assess the capability of five state-of-the-art VLMs to predict the underlying structure and semantics of a scene.
For each model, we report its performance when recognizing the property of an object (RQ1), evaluate its robustness to variations in the object positioning (RQ2), and measure its ability to identify relations among objects (RQ3).
Finally, we compare the model's performance with that of humans, and the human performance with the ground truth of the stimuli.
Following preliminary experiments on image and object sizes (see Appendix \ref{sec:app-img-obj-size}), we select the best setting, i.e., \textit{regular} size objects with $672\times672$ images for elementary objects, and $1344\times1344$ for COCO objects.

\begin{table}[!t]
    \centering
    \small
    \begin{adjustbox}{width=\columnwidth}
        \begin{tabularx}{1.05\columnwidth}{lrrrrrr}
            \toprule
            \multirow{2}{*}{\textbf{M\scriptsize{odel}}} & \multicolumn{6}{c}{\textbf{C\scriptsize{olor}}} \\
            \cmidrule{2-7}
            & \texttt{R} & \texttt{G} & \texttt{B} & \texttt{Y} & \texttt{M} & \texttt{C} \\
            \midrule
            \textit{LLaVA-NeXT 7B}  
            &	100	&	87	&	83	&	95	&	99	&	56\\
            \midrule
            \textit{LLaVA-NeXT 13B}	            
            &	88	&	86	&	74	&	98	&	74	&	2\\
            \midrule
            \textit{Molmo-O 7B}
            &	98	&	98	&	96	&	99	&	98	&	96\\
            \midrule
            \textit{Qwen2-VL 7B}
            &	100	&	100	&	99	&	100	&	97	&	100\\
            \midrule
            \textit{CLIP}
            &	100	&	99	&	88    &	100	&	100	&	82\\
            \bottomrule
        \end{tabularx}
    \end{adjustbox}
    \caption{F1-Score (\%) of each model when considering visual scenes containing a single object and querying about its \textit{color}. There are six possible colors: red (\texttt{R}), green (\texttt{G}), blue (\texttt{B}), yellow (\texttt{Y}), magenta (\texttt{M}), and cyan (\texttt{C}). Results are based on $672 \times 672$ images.}
    \label{tab:f1-color}
\end{table}

\begin{figure*}[t]
    \centering
    \includegraphics[width=0.75\linewidth]{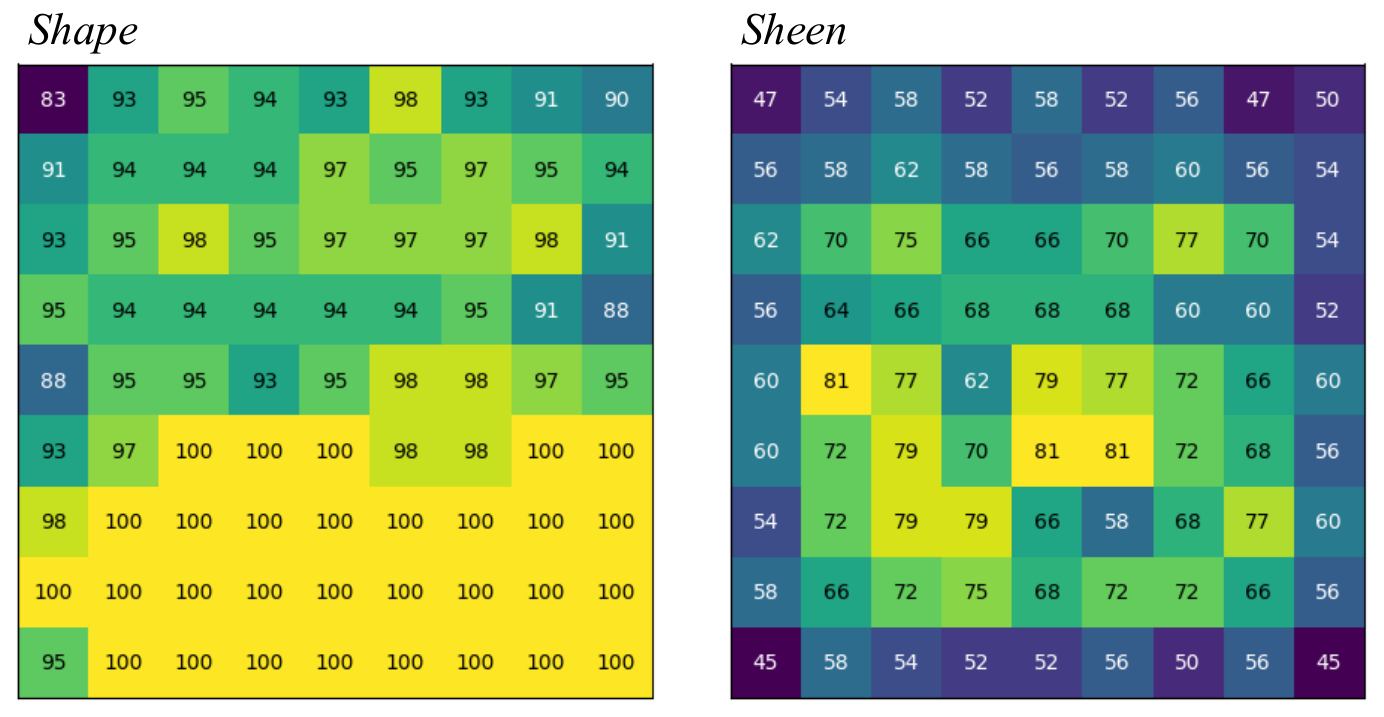}
    \caption{
        Accuracy (\%) of \llavas 13B in each cell of our $9 \times 9$ world when queried about the \textit{Shape} \& \textit{Sheen} of a Single Object. Results are based on $672\times672$ images.
    }
    \label{fig:shape-sheen-given-pos}
\end{figure*}

\subsection{RQ1: Can VLMs accurately recognize basic object properties?}
We report results on VLMs' recognition of basic properties and absolute positions when analyzing a single, elementary object.
To consolidate our findings and avoid overestimating VLMs performance, we extend our study by testing these models on real-world objects.

\textbf{Single Object}
\label{sec:single-obj}
Table \ref{tab:rq1-acc} shows the results of the Single Object experiment (Section \ref{par:single-obj}) considering elementary objects.
Since all classes in our dataset are balanced, we report performance in terms of accuracy.

Among our set of properties, \textit{shape} was the easiest to recognize, with models obtaining at least 95\%.
Conversely, VLMs achieved the worst performance when predicting the \textit{sheen} of the object, with the best accuracy reaching 64\%.
Regarding the \textit{color}, the \llavas models obtained the worst results, with the smaller 7B model being 12\% more accurate than its larger 13B counterpart, suggesting that scaling the text decoder does not always improve performance.
When considering colors individually (Table \ref{tab:f1-color}), both models showed lower performance on green, blue, and especially cyan (with the smaller 7B model achieving 56\% F1, while the larger 13B model only 2\%).
This can be partly explained by the performance of their vision encoder, CLIP, which showed lower performance on blue and cyan.
Nevertheless, despite using \clip, \molmos showed no relevant difference in performance on these colors, obtaining almost perfect accuracy (along with \qwen).
Similarly, \molmos and \qwens achieved the best results when querying about the \textit{absolute position} of an object, while \llavas models achieved around 15\% less accuracy.
When comparing the performance of the other VLMs with \clip, no significant differences can be observed when querying for the properties (i.e., \textit{shape}, \textit{color}, \textit{sheen}) of an object.
However, while \clips obtained close to random performance when predicting the \textit{absolute position} of an object (w.r.t. the background), \llavas and \molmos showed considerable improvements (up to 35\%), suggesting that the LLMs used as text decoders may have a positive impact on visual tasks.
However, the different fine-tuning data may also be responsible for the improvements.

\textbf{Single Object w. COCO}
\label{sec:single-obj-coco}
To extend our findings to real-world scenes, we experiment with objects extracted from the COCO dataset.
We evaluate the models when predicting the \textit{category} and the \textit{absolute position} of the object, and report the results in Table \ref{tab:coco-rq1-acc}.
When querying about the object \textit{category}, all VLMs showed higher accuracy than CLIP, suggesting that fine-tuning and the additional text decoder (LLM) can benefit visual-language tasks.
Similarly to the synthetic objects, increasing the size of the text decoder (from 7B to 13B) does not always improve performance.
Regarding the \textit{absolute position}, performance with COCO objects follows an analogous trend w.r.t. synthetic objects (with CLIP performing close to random).

\begin{figure*}
    \centering
    \includegraphics[width=0.955\linewidth]{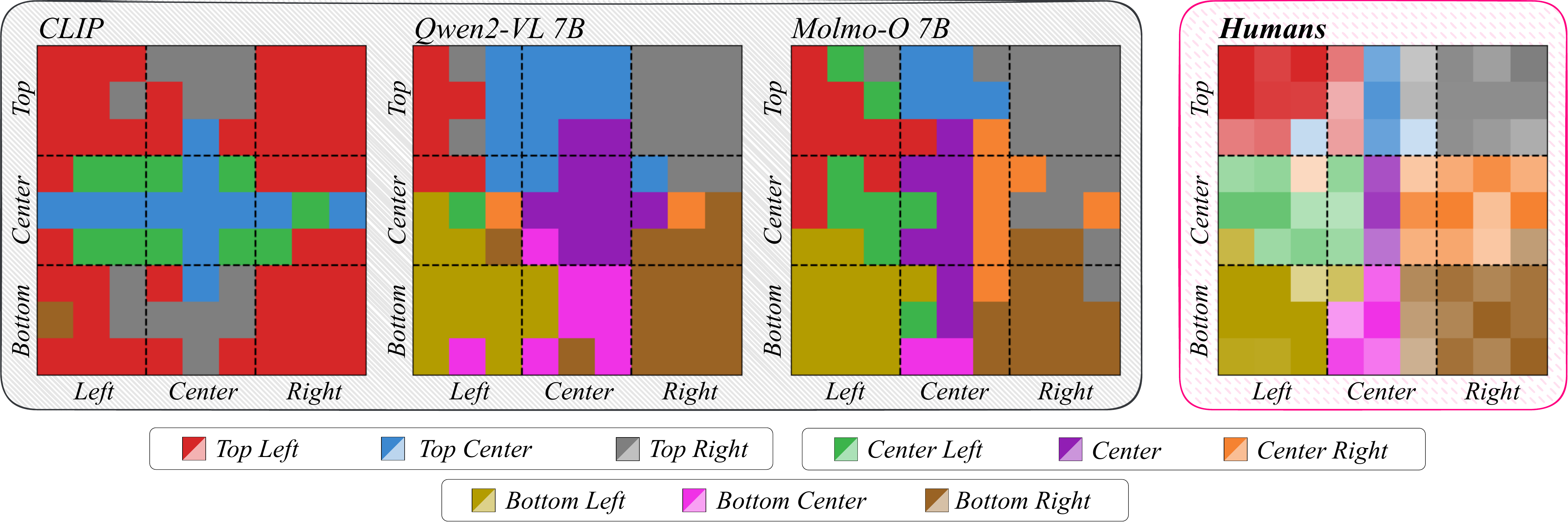}
    \caption{
        \textit{"Where is the yellow star?"} - Responses of CLIP, Qwen, Molmo-O 7B, and Humans when asking a closed-ended question about the position of a yellow star on a black background.
        The question was asked by placing the object in all cells of a $9\times9$ grid.
        Since we obtained multiple human annotations for the same stimulus, we report the majority vote.
        Dashed lines delimit the ground truth sections, and colors indicate the response for each cell. For Humans, the colors fade to white to represent the decrease in agreement (\% votes for the majority class).
        Results are based on $672\times672$ images.
    }
    \label{fig:common-pos}
\end{figure*}

\subsection{RQ2: Is VLM performance robust to variations in object positioning?}
Since understanding requires consistent performance across varying conditions, we study whether changes in object position within the visual scene affect the recognition of object properties.
We then investigate how VLMs associate natural language position (e.g., \textit{top left}) with specific regions of a scene, and compare their predictions to those of humans using identical stimuli.

\textbf{Effect of Object Position on Accuracy}
We measure the accuracy in each cell of our $9\times9$ world and report additional results in Appendix \ref{sec:app-position}.
In all Single Object experiments, we find that accuracy is not uniform over all cells, but varies considerably when changing the position of the object.
As an example, we show the results of \llavas 13B on \textit{shape} and \textit{sheen} in Figure \ref{fig:shape-sheen-given-pos}. 
Regarding the \textit{shape} of an object, \llavas 13B performed worst in the top corners (83\% and 90\%), but achieved 100\% accuracy in almost every cell of the bottom part of the scene, suggesting the tendency to look at the last visual tokens.
When querying about the \textit{sheen}, the overall performance of \llavas 13B was poor (64\% as shown in Table \ref{tab:rq1-acc}).
However, Figure \ref{fig:shape-sheen-given-pos} shows that its performance on sheen reached 81\% near the center, while it dropped to a minimum of 45\% towards the corners.

Regarding the \textit{absolute position} of the object, models show higher performance in the corners and the center.
Because these models are mostly trained on image-caption pairs, their definition of \textit{top left} could refer to a different part of the scene.
Since we arbitrarily assigned each cell to a particular absolute position (see Single Object in \ref{par:single-obj}), we report the models' position assignment for each cell (Figure \ref{fig:common-pos}) when considering the object that models recognized best (\textit{yellow star}).
Although each VLM tends to use a different position assignment, \qwens and \molmos assign the corners and the center to the correct position.
In particular, they obtained almost perfect accuracy on the \textit{top right}, \textit{bottom right}, and \textit{bottom left} sections of the scene, indicating that there may be a bias in their training data.
Moreover, these models rarely invert top with bottom or left with right, suggesting that they have some understanding of position.
The only exception is \clip, which showed close to random performance and assigned \textit{top center} to the central section of the scene and \textit{top left} to most of the other cells.

\textbf{Human Evaluation: Absolute Position}
To get more insights about our arbitrary position assignment, we conducted a human evaluation with a subset of input stimuli used with VLMs.
For the evaluation, we selected all 81 visual scenes containing one \textit{yellow star} (i.e., one for each cell of our $9\times9$ world), given that it was the \textit{shape}-\textit{color} combination with the highest average F1 across all models.
Similarly to what we did for the models, we asked human annotators ``Where is the yellow star?'' while providing the set of possible answers.
We report the guidelines and the user interface for the annotation in Appendix \ref{sec:human-eval}.
For the annotation, we recruited 124 English-speaking Amazon Mechanical Turkers\footnote{\url{https://www.mturk.com/}} and assigned 10 stimuli to each.
Of these participants, 91 were approved after quality control and compensated with \$2.00\footnote{This corresponds to \$24.00/hour, given that the task took 5 minutes on average}.
Each stimulus was annotated 8 times and the inter-annotator agreement as measured by Fleiss' $\kappa$ \cite{fleiss1971measuring} was 0.61 (substantial agreement).

Based on the annotation results, humans achieved an accuracy of 73\%, higher than all the models.
When considering majority voting to determine the \textit{absolute position} of the object (see Figure \ref{fig:common-pos}), humans separate the vertical component in three equal bands, granting them an almost perfect performance on the vertical component (\textit{top}, \textit{center}, \textit{bottom}).
Instead, for the horizontal component (\textit{left}, \textit{center}, \textit{right}), humans tend to shrink the area for the center to only the central row of positions, leaving more to left and right.
Regarding their confidence (i.e., the ratio between the most voted label and total votes), humans show higher agreement near the corners and the center, while lower near borders and around the center.
This suggests that humans are capable of locating a position in the scene (e.g., \textit{center}, or \textit{top left}), but are biased towards \textit{left} and \textit{right}.

When comparing with the models, \molmos is the only model assigning a narrow set of cells in the center for the horizontal component, showing similar performance to humans (possibly due to human-annotated positions in its training data \cite{deitke2024molmo}).
On the other hand, \qwens assigned a higher number of cells to both \textit{top-center} and \textit{bottom-center}, resembling our position assignment.

\subsection{RQ3: Can VLMs identify basic relations among objects?}

\begin{table}[t]
    \centering
    \small
    \begin{tabularx}{0.88\columnwidth}{lrrr}
        \toprule
        \multirow{2}{*}{\textbf{M\scriptsize{odel}}} & \multicolumn{3}{c}{\textbf{R\scriptsize{elative}}} \\
        \cmidrule{2-4}
        & \texttt{P\scriptsize{osition}} & \texttt{D\scriptsize{istance}} & \texttt{S\scriptsize{ize}} \\
        \midrule
        \textit{Random Baseline}
        &	13	&	50	&	33\\
        \midrule
        \textit{LLaVA-NeXT 7B}
        &	24	&	54	&	30\\
        \midrule
        \textit{LLaVA-NeXT 13B}
        &	38	&	59	&	33\\
        \midrule
        \textit{Molmo-O 7B}
        &	24	&	76	&	30\\
        \midrule
        \textit{Qwen2-VL 7B}
        &	\textbf{46}	&	\textbf{83}	&	\textbf{54}\\
        \midrule
        \textit{CLIP}
        &	20	&	51	&	49\\
        \bottomrule
    \end{tabularx}
    \caption{Accuracy (\%) of each model when considering visual scenes containing different objects and querying about the \textit{relative position}, \textit{relative distance}, and \textit{relative size} of one object w.r.t. the others. Results are based on $672 \times 672$ images.}
    \label{tab:acc-rq2}
\end{table}
Table \ref{tab:acc-rq2} shows the accuracy when predicting the \textit{relative position}, \textit{relative distance}, and \textit{relative size} among the objects.
Regarding the \textit{relative position}, all VLMs achieved a higher accuracy than \clips (20\%), with \qwens obtaining the highest performance (46\%).
Results show that relations about objects on the same row/column (i.e., \textit{directly above}, \textit{directly left}, \textit{directly right}, \textit{directly below}) were harder to predict, with \qwens achieving an average of 18.5\% F1.
Additionally, \clips achieved 33\% F1 on \textit{above left} but 0\% on all other relations, suggesting the presence of a strong bias.
Additional results can be found in Appendix \ref{sec:app-rel-pos}.

When queried about the \textit{relative distance} of an object (i.e., identifying the closest object), \clips and \llavas models (7B \& 13B) achieved performance close to random.
As shown in Table \ref{tab:f1-rel-dis}, this is partly due to the failure of \llavas models to detect the closest object to a \textit{triangle}.
However, when predicting the \textit{shape} of a Single Object, \llavas models obtained almost a perfect F1-score on \textit{triangle} ($\geq96\%$ F1).
CLIP shows a similar problem with the \textit{circle}, which, although it was the shape with the lowest performance, obtained an F1 score of $91\%$.
These findings suggest that, despite being able to recognize the \textit{shape} of an object, some VLMs are unable to use this property to refer to an object, making their performance task-dependent.

\begin{table}[t]
    \centering
    \small
    \begin{tabularx}{0.84\columnwidth}{lrrr}
        \toprule
        \multirow{2}{*}{\textbf{M\scriptsize{odel}}} & \multicolumn{3}{c}{\textbf{R\scriptsize{elative} \small{D}\scriptsize{istance}}} \\
        \cmidrule{2-4}
        & \texttt{C\scriptsize{ircle}} & \texttt{S\scriptsize{tar}} & \texttt{T\scriptsize{riangle}} \\
        \midrule
        \textit{LLaVA-NeXT 7B}
        &	62	&	64	&	14\\
        \midrule
        \textit{LLaVA-NeXT 13B}
        &	67	&	71	&	10\\
        \midrule
        \textit{Molmo-O 7B}
        &	76	&	79	&	71\\
        \midrule
        \textit{Qwen2-VL 7B}
        &	83	&	85	&	81\\
        \midrule
        \textit{CLIP}
        &	0	&	55	&	65\\
        \bottomrule
    \end{tabularx}
    \caption{F1-Score (\%) of each model when considering visual scenes containing multiple objects and querying about the \textit{relative distance} of one object w.r.t. other two objects. Results are based on $672 \times 672$ images.}
    \label{tab:f1-rel-dis}
\end{table}

\textit{Relative size} proved to be the hardest relation to predict since only \qwens and \clips achieved better than random performance.
As reported in Table \ref{tab:f1-rel-size}, the other VLMs never predicted \textit{same} correctly (while \qwens achieved 47\% F1 and \clips 66\% F1).
Despite its performance, \clips showed poor results on \textit{smaller} and \textit{larger} ($\leq17\%$), indicating a bias for \textit{same}.

\begin{table}[t]
    \centering
    \small
    \begin{tabularx}{0.82\columnwidth}{lrrr}
        \toprule
        \multirow{2}{*}{\textbf{M\scriptsize{odel}}} & \multicolumn{3}{c}{\textbf{R\scriptsize{elative Size}}} \\
        \cmidrule{2-4}
        & \texttt{S\scriptsize{maller}} & \texttt{S\scriptsize{ame}} & \texttt{L\scriptsize{arger}} \\
        \midrule
        \textit{LLaVA-NeXT 7B}
        &	41	&	0	&	38\\
        \midrule
        \textit{LLaVA-NeXT 13B}
        &	45	&	0	&	42\\
        \midrule
        \textit{Molmo-O 7B}
        &	42	&	0	&	33\\
        \midrule
        \textit{Qwen2-VL 7B}
        &	57	&	47	&	61\\
        \midrule
        \textit{CLIP}
        &	17	&	66	&	0\\
        \bottomrule
    \end{tabularx}
    \caption{F1-Score (\%) of each model when considering visual scenes containing multiple objects and querying about the \textit{relative size} of one object w.r.t. another. Results are based on $672 \times 672$ images.}
    \label{tab:f1-rel-size}
\end{table}

\section{Conclusion}
In this work, we studied whether state-of-the-art VLMs understand the underlying structure and semantics of a visual scene.
To respond to the lack of standardized and systematic evaluation, we introduce CIVET, a framework to systematically assess VLM's understanding via controlled stimuli.
Our study reveals that 1) VLMs are only capable of recognizing certain properties, 2) their performance heavily depends on the position of the object, and 3) they struggle to identify basic relations among objects.
Moreover, a comparative analysis with human annotators shows that VLMs fall short of human-level accuracy. 
Our findings indicate that VLMs have limited understanding, which limits their generalization in learning.
We encourage further community engagement to extend CIVET and promote novel training paradigms that are pedagogical rather than utility driven, i.e. fine-tuning for downstream tasks.

\section*{Limitiations}
Due to the limited computational resources, we could not experiment with larger models, limiting the results on the effect of model size to 7B and 13B models.
Regarding the effect of the objects' size, further study is needed, as we only considered two variants.
Furthermore, differences in the sets of crowd workers may result in variations in the human evaluation.

\section*{Ethical Statement}
The engagement of crowd-workers for human evaluation does not introduce any ethical concern since the task solely consisted of annotating the position of a yellow star on a back background, which has a low cognitive load.


\bibliography{acl_latex}

\clearpage
\appendix

\section{Appendix}
\label{sec:appendix}

\subsection{Experimental Details}
\label{sec:exp-details}
Most experiments were executed using one NVIDIA A100 with 40 GiB. The only exception was Qwen2-VL-7B-Instruct, which required one NVIDIA A100 with 80 GiB when considering images of size $1344\times1344$.
In all experiments, we used greedy generation to generate the answers to our closed-ended questions.
Regarding the models, we considered the following HuggingFace checkpoints:
\begin{enumerate}
    \item \normalsize \llavas 7B, \small\url{https://huggingface.co/llava-hf/llava-v1.6-vicuna-7b-hf}
    \item \normalsize \llavas 13B, \small\url{https://huggingface.co/llava-hf/llava-v1.6-vicuna-13b-hf}
    \item \normalsize \molmos 7B-O, \small\url{https://huggingface.co/allenai/Molmo-7B-O-0924}
    \item \normalsize\qwen-7B-Instruct, \small\url{https://huggingface.co/Qwen/Qwen2-VL-7B-Instruct}
    \item \normalsize \clips ViT-L/14-336px, \small\url{https://huggingface.co/openai/clip-vit-large-patch14}
\end{enumerate}

\subsection{Answers Length and "Other" values}
\label{subsec:answ-length}

In a preliminary experiment on Single Object (see Section \ref{sec:dataset}), we noticed that only \molmos answered with one token while the other VLMs answered the questions with 8 to 15 tokens (tokenized with NLTK\footnote{\url{https://www.nltk.org/}}).
Because of this, only manual checking would have been appropriate, as exact matching would have resulted in underestimating the performance of more verbose models.
For this reason, we prepended the instruction \textit{"Answer with as few words as possible."} as an attempt to condition the model to generate only the property values as the answer.
Table \ref{tab:ans-len-instr} shows this had the desired effect of reducing the number of tokens to 1 ($\pm0.5$ for \qwens only) for the property questions, and 2 tokens ($\pm 0.5$ at most) for the position questions.

Since we used greedy generation, we also measured how often the models responded with "other" answers: either something outside the provided set or multiple options from the set.
When computing accuracy, we considered "other" answers as mistakes.
As shown in Table \ref{tab:others}, \molmos was the only model that never generated "other" answers. 
Instead, \qwens shows a small percentage of "other" answers (below 2\%) regardless of the question, while the two \llavas models only generate them when querying about color (under 1\% for 7B, and under 6.8\% for 13B).
Additionally, adding the instruction for shorter answers reduced the frequency of "other" answers.
We report the full tables for the answer lengths in Table \ref{tab:ans-len-instr} and the percentage of "other" answers in Table \ref{tab:others}.

\subsection{Effect of Image and Object Size}
\label{sec:app-img-obj-size}
As high-resolution images are known to increase the performance of VLMs \cite{NEURIPS2024_a0303731, karamcheti2024prismatic}, we performed additional experiments to understand how different image and object sizes could affect the model performance.
We experiment with three image sizes ($336\times336$, $672\times672$, and $1344\times1344$) and two object sizes (\textit{regular}, and \textit{small}) to understand how these parameters affect the models' performance when predicting object properties and absolute position.

Table \ref{tab:rq1-acc-full} shows the accuracy of each model when considering the three image sizes.
Increasing the image size affects \clips performance only negligibly (variations are likely to be attributed to the resize operation).
On the other hand, VLMs achieve higher accuracy with larger images, suggesting that encoding multiple high-resolution patches helps capture more details.
When considering synthetic objects, increasing the image size to $672\times672$ leads to a higher performance. 
However, further increasing the image size to $1344\times1344$ does not provide any significant improvement.
When considering COCO objects, \clip, \llava, and \molmos show a similar trend.
Instead, \qwens is the only model improving accuracy (97\% on \textit{category}), suggesting that encoding the whole image with no resizing is more advantageous than encoding multiple high-resolution patches (\llava, and \molmo).
Regarding the size of the object, using \textit{small} objects leads to a drop in performance in most cases for all models. 

Table \ref{tab:rq1-delta} shows the difference in accuracy between \textit{regular} (i.e., Table \ref{tab:rq1-acc-full}) and \textit{small} objects (i.e., resized by $\frac{1}{4}$). Similarly, Table \ref{tab:rq1-delta-coco} shows the difference in accuracy between \textit{regular} and \textit{small} objects when considering COCO Objects. In both cases, most model performance is higher when considering \textit{regular} objects.

As image and object size affect performance, for the rest of the experiments we report only the results in the best setting, i.e., \textit{regular} size objects with $672\times672$ images for synthetic objects, and $1344\times1344$ for COCO objects.

\subsection{Single Object}
We provide the F1-Score for the remaining Single Object experiments when considering images with $672\times672$ pixels.

Tables \ref{tab:f1-shape}, \ref{tab:f1-sheen}, and \ref{tab:f1-abs} show the F1-Score of each model when considering only one object and querying about its \textit{shape}, \textit{sheen}, and \textit{position} (w.r.t. the background), respectively.

\subsection{Effect of Object Position on Accuracy}
\label{sec:app-position}
\begin{figure*}[b]
    \centering
    \includegraphics[width=0.85\linewidth]{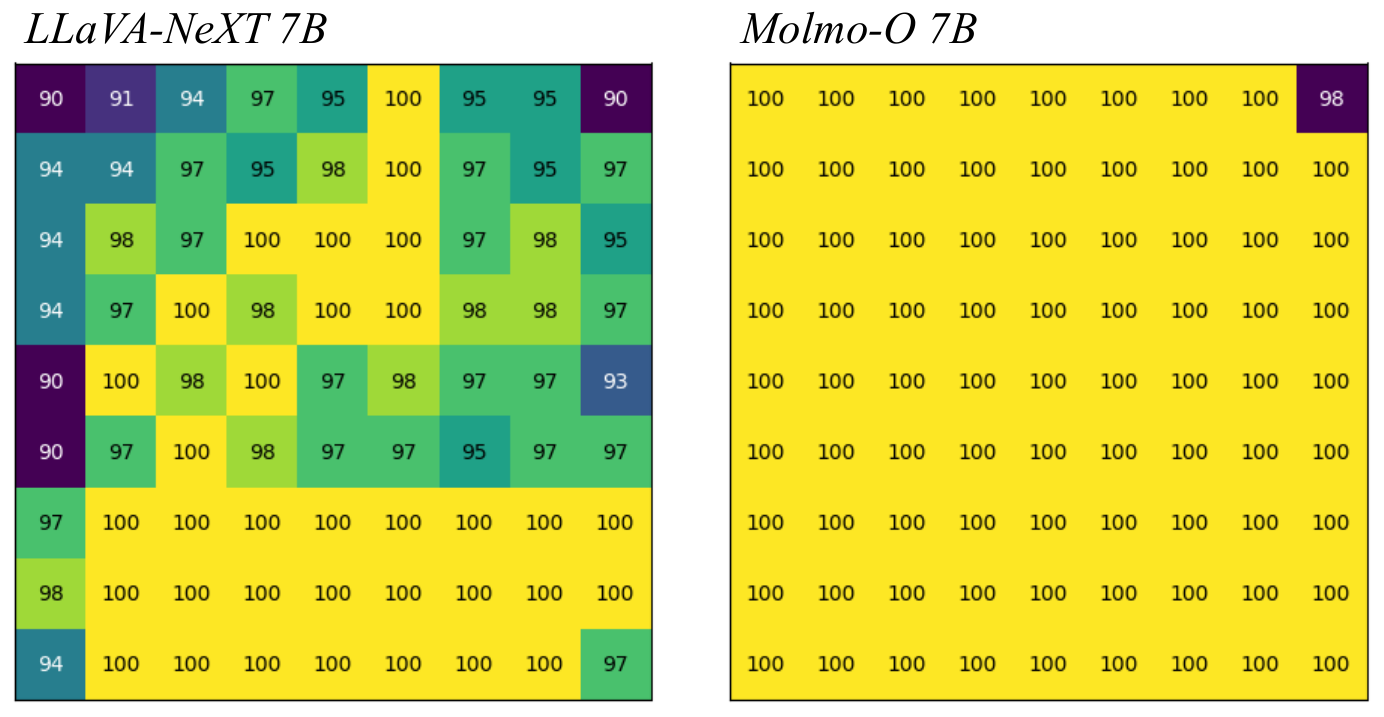}
    \caption{
        Accuracy (\%) of \llavas 7B \& \molmos 7B in each cell of our $9 \times 9$ world when queried about the \textit{shape} of a Single Object. Results are based on $672\times672$ images.
    }
    \label{fig:shape-given-pos}
\end{figure*}

We report additional findings when investigating the accuracy of VLMs in each cell of our $9 \times 9$ world.
Figure \ref{fig:shape-given-pos} shows the results for \llavas 7B and \molmos when asking about the shape of an object.
Similarly to its larger counterpart (shown in Figure \ref{fig:shape-sheen-given-pos}), \llavas 7B performed worse in the top corners, obtaining almost perfect accuracy in the bottom part of the scene and showing the same tendency to look at the last visual tokens.
Since both \llavas models suffer from this bias, it may be related to the data used to train the projection layer.
On the other hand, \molmo, whose accuracy was 100\% when recognizing the shape of an object, is the only model showing almost perfect accuracy across the whole scene (except for the cell in the upper right corner).

\begin{figure*}[b]
    \centering
    \includegraphics[width=0.85\linewidth]{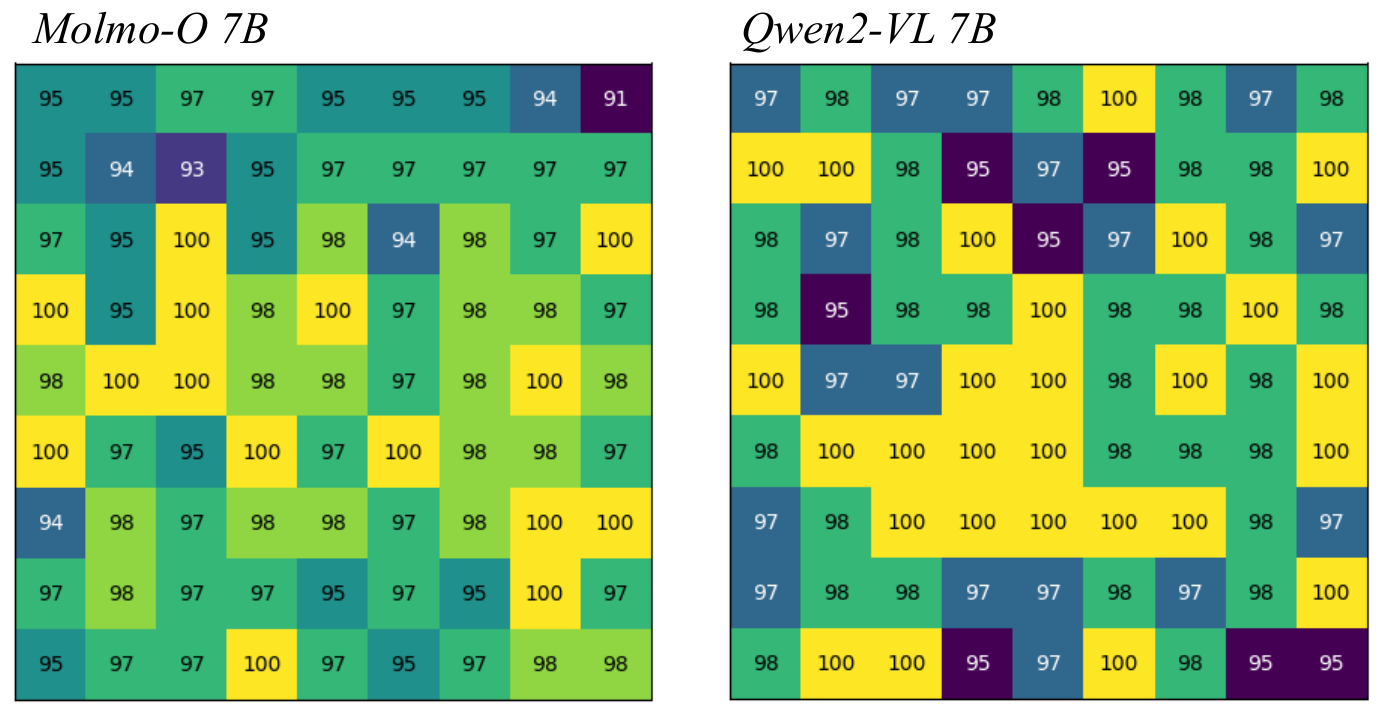}
    \caption{
        Accuracy (\%) of \molmos 7B \& \qwens 7B in each cell of our $9 \times 9$ world when queried about the \textit{color} of a Single Object. Results are based on $672\times672$ images.
    }
    \label{fig:color-given-pos}
\end{figure*}

When queried about the \textit{color} of a Single Object, \molmos and \qwens obtained 98\% and 99\% accuracy, respectively (see Table \ref{tab:rq1-acc}).
However, when looking at Figure \ref{fig:color-given-pos} it is possible to notice how performance depends on the position of the object.
Similarly to \textit{shape} (Figure \ref{fig:shape-given-pos}), \molmos obtained the worst accuracy when the object was placed in the top right corner of the scene, possibly indicating the presence of a bias.
In general, cells with higher and lower accuracy are randomly spread across the whole scene.
This also happens for \qwen, with the only difference that cells with the lowest accuracy are more present in the last row (i.e., bottom part) of the scene.

\begin{figure}[b]
    \centering
    \includegraphics[width=0.95\columnwidth]{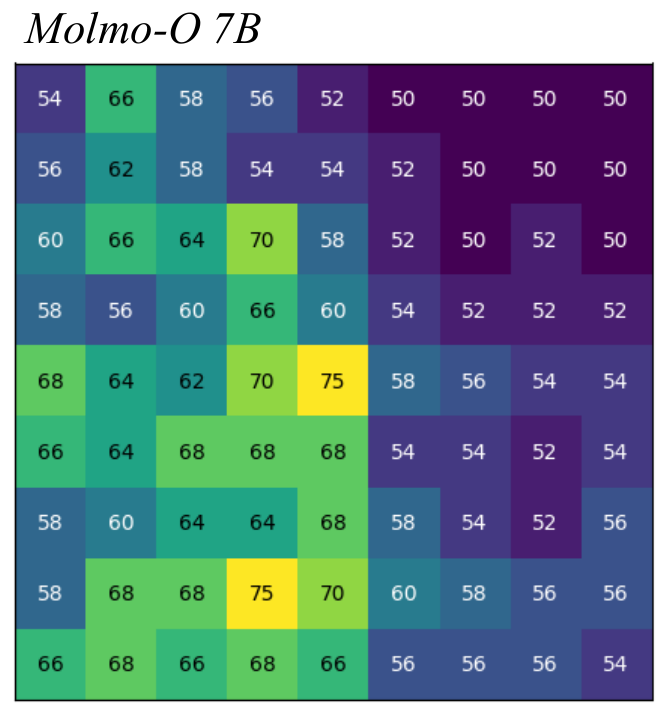}
    \caption{
        Accuracy (\%) of \molmos 7B in each cell of our $9 \times 9$ world when queried about the \textit{sheen} of a Single Object. Results are based on $672\times672$ images.
    }
    \label{fig:color-given-sheen}
\end{figure}

Figure \ref{fig:color-given-sheen} shows \molmos accuracy when asking about the \textit{sheen} of an object when placed in different cells of our world. Similarly to \llavas 13B (see Figure \ref{fig:shape-sheen-given-pos}), \molmos accuracy is higher towards the cells in the center. However, performance is on average worse for all the cells in the right section of the scene (especially in the top right).

\subsection{Relative Position}
\label{sec:app-rel-pos}
Table \ref{tab:f1-rel-pos} shows the F1-Score of each model when considering visual scenes containing two objects and querying about the \textit{relative position} of one object w.r.t. the other.

\subsection{Experiments on COCO}
Table \ref{tab:coco-rq1-acc-full} shows the accuracy of each model when considering different objects from the COCO dataset (i.e., zebra, giraffe, and elephant).
Additionally, Table \ref{tab:f1-coco-cat} shows the F1-Score of each model when considering visual scenes containing different objects from COCO (i.e., zebra, giraffe, and elephant) and querying about their \textit{category}.

\subsection{Human Evaluation}
\label{sec:human-eval}
\begin{figure*}[t!]
    \centering
    \includegraphics[width=\linewidth]{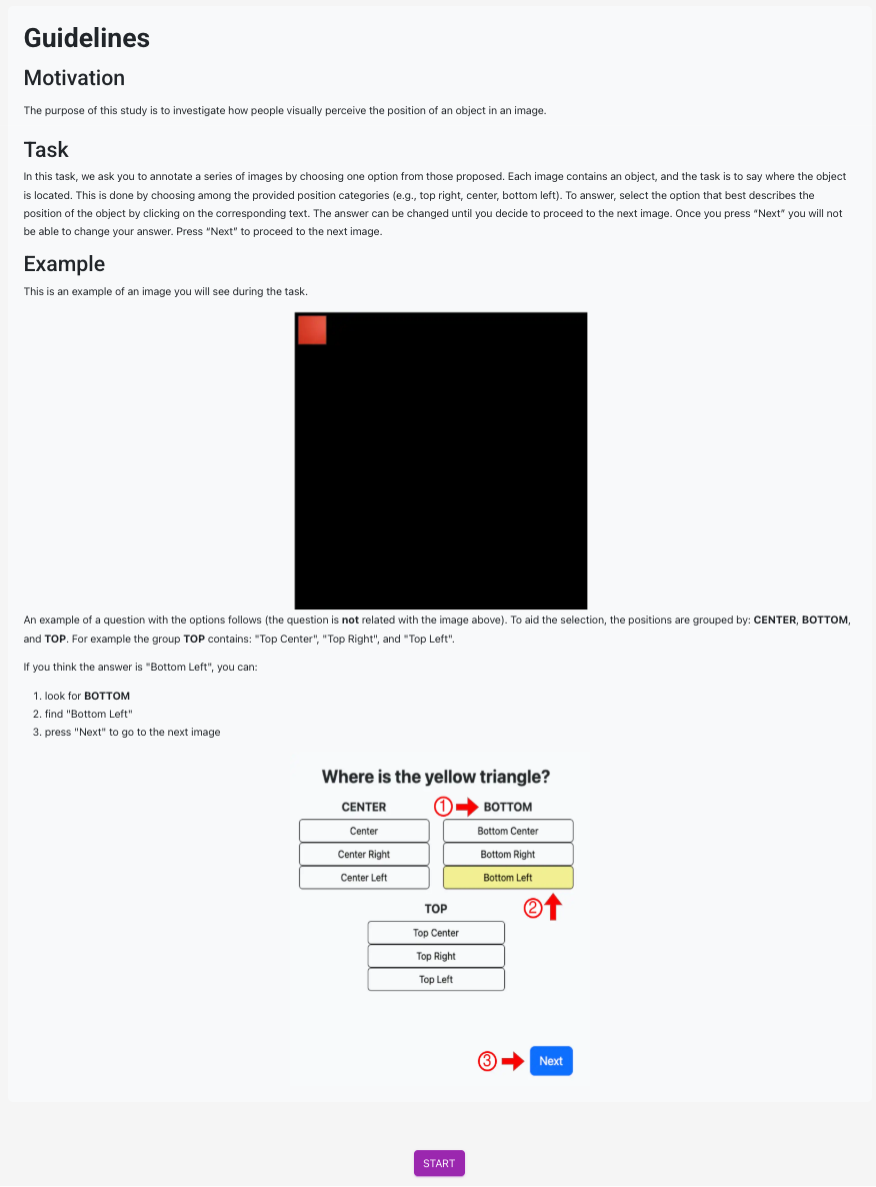}
    \caption{Guidelines for the proposed human evaluation task.}
    \label{fig:guidelines}
\end{figure*}

\begin{figure*}[t!]
    \centering
    \includegraphics[width=\linewidth]{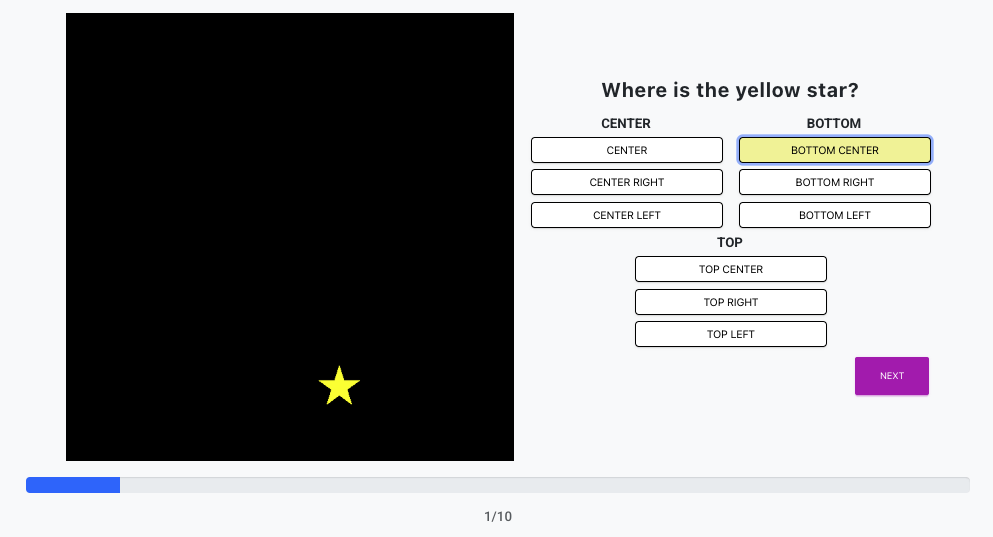}
    \caption{User interface for the proposed human evaluation task.}
    \label{fig:ui}
\end{figure*}

We report the guidelines provided to the human annotators in Figure \ref{fig:guidelines} and the user interface for the annotation task in Figure \ref{fig:ui}.

\begin{table*}[t]
    \small
    \centering
    \begin{tabularx}{0.57\textwidth}{lrrrrrrr}
        \toprule
        \textbf{M\scriptsize{odel}} & \textbf{I\scriptsize{mage} \small{S}\scriptsize{ize}} & \textbf{S\scriptsize{hape}} & \textbf{C\scriptsize{olor}} & \textbf{S\scriptsize{heen}} & \textbf{P\scriptsize{osition}} \\
        \midrule
        \multirow{3}{*}{\textit{LLaVA-NeXT 7B}}  
        &   336	    &	\textcolor[HTML]{228B22}{$\uparrow$}7	&	\textcolor[HTML]{228B22}{$\uparrow$}2	&	0	&	\textcolor[HTML]{228B22}{$\uparrow$}1\\
        &   672	    &	\textcolor[HTML]{228B22}{$\uparrow$}6	&	0	&	\textcolor{red}{$\downarrow$}1	&	0\\
        &   1344	&	\textcolor[HTML]{228B22}{$\uparrow$}5	&	\textcolor{red}{$\downarrow$}1	&	0	&	0\\
        \midrule
        \multirow{3}{*}{\textit{LLaVA-NeXT 13B}}	            
        &   336	    &	\textcolor[HTML]{228B22}{$\uparrow$}6	&	0	&	\textcolor[HTML]{228B22}{$\uparrow$}3	&	\textcolor[HTML]{228B22}{$\uparrow$}1\\
        &   672	    &	\textcolor[HTML]{228B22}{$\uparrow$}4	&	0	&	\textcolor[HTML]{228B22}{$\uparrow$}6	&	\textcolor[HTML]{228B22}{$\uparrow$}2\\
        &   1344	&	\textcolor[HTML]{228B22}{$\uparrow$}3	&	0	&	\textcolor[HTML]{228B22}{$\uparrow$}12	&	\textcolor[HTML]{228B22}{$\uparrow$}2\\
        \midrule
        \multirow{3}{*}{\textit{Molmo-O 7B}}
        &   336	    &	\textcolor[HTML]{228B22}{$\uparrow$}5	&	\textcolor[HTML]{228B22}{$\uparrow$}1	&	0	&	\textcolor[HTML]{228B22}{$\uparrow$}4\\
        &   672	    &	0	&	\textcolor[HTML]{228B22}{$\uparrow$}2	&	\textcolor[HTML]{228B22}{$\uparrow$}1	&	\textcolor[HTML]{228B22}{$\uparrow$}2\\
        &   1344	&	0	&	\textcolor[HTML]{228B22}{$\uparrow$}1	&	\textcolor[HTML]{228B22}{$\uparrow$}1	&	\textcolor[HTML]{228B22}{$\uparrow$}2\\
        \midrule
        \multirow{3}{*}{\textit{Qwen2-VL 7B}}
        &   336	    &	\textcolor[HTML]{228B22}{$\uparrow$}9	&	\textcolor[HTML]{228B22}{$\uparrow$}2	&	\textcolor[HTML]{228B22}{$\uparrow$}3	&	\textcolor[HTML]{228B22}{$\uparrow$}1\\
        &   672	    &	\textcolor[HTML]{228B22}{$\uparrow$}1	&	\textcolor[HTML]{228B22}{$\uparrow$}1	&	\textcolor[HTML]{228B22}{$\uparrow$}5	&	\textcolor[HTML]{228B22}{$\uparrow$}1\\
        &   1344	&	\textcolor[HTML]{228B22}{$\uparrow$}1	&	0	&	\textcolor[HTML]{228B22}{$\uparrow$}4	&	\textcolor[HTML]{228B22}{$\uparrow$}1\\
        \midrule
        \multirow{3}{*}{\textit{CLIP}}
        &   336	    &	\textcolor[HTML]{228B22}{$\uparrow$}4	&	\textcolor{red}{$\downarrow$}2	&	0	&	\textcolor{red}{$\downarrow$}1\\
        &   672	    &	0	&	0	&	\textcolor{red}{$\downarrow$}1	&	0\\
        &   1344	&	\textcolor[HTML]{228B22}{$\uparrow$}1	&	\textcolor[HTML]{228B22}{$\uparrow$}1	&	\textcolor{red}{$\downarrow$}1	&	0\\
        \bottomrule
    \end{tabularx}
    \caption{Difference in accuracy between \textit{Regular} (i.e., Table \ref{tab:rq1-acc-full}) and \textit{small} objects (i.e., resized by $\frac{1}{4}$)}
    \label{tab:rq1-delta}
\end{table*}

\begin{table}[t]
    \small
    \centering
    \begin{tabularx}{0.95\columnwidth}{lrrr}
        \toprule
        \textbf{M\scriptsize{odel}} & \textbf{I\scriptsize{mage} \small{S}\scriptsize{ize}} & \textbf{C\scriptsize{ategory}} & \textbf{P\scriptsize{osition}} \\
        \midrule
        \multirow{3}{*}{\textit{LLaVA-NeXT 7B}}  
        &   336	    &		\textcolor[HTML]{228B22}{$\uparrow$}15	&		\textcolor[HTML]{228B22}{$\uparrow$}2\\
        &   672	    &		\textcolor[HTML]{228B22}{$\uparrow$}50	&		\textcolor[HTML]{228B22}{$\uparrow$}5\\
        &   1344	&		\textcolor[HTML]{228B22}{$\uparrow$}47	&		\textcolor[HTML]{228B22}{$\uparrow$}11\\
        \midrule
        \multirow{3}{*}{\textit{LLaVA-NeXT 13B}}
        &   336	    &		\textcolor[HTML]{228B22}{$\uparrow$}26	&		\textcolor[HTML]{228B22}{$\uparrow$}5\\
        &   672	    &		\textcolor[HTML]{228B22}{$\uparrow$}28	&		\textcolor[HTML]{228B22}{$\uparrow$}1\\
        &   1344	&		\textcolor[HTML]{228B22}{$\uparrow$}15	&		\textcolor[HTML]{228B22}{$\uparrow$}12\\
        \midrule
        \multirow{3}{*}{\textit{Molmo-O 7B}}
        &   336	    &		\textcolor[HTML]{228B22}{$\uparrow$}1	&		\textcolor[HTML]{228B22}{$\uparrow$}7\\
        &   672	    &		\textcolor[HTML]{228B22}{$\uparrow$}34	&		\textcolor[HTML]{228B22}{$\uparrow$}14\\
        &   1344	&		\textcolor[HTML]{228B22}{$\uparrow$}37	&		\textcolor[HTML]{228B22}{$\uparrow$}7\\
        \midrule
        \multirow{3}{*}{\textit{Qwen2-VL 7B}}
        &   336	    &		\textcolor[HTML]{228B22}{$\uparrow$}32	&		\textcolor[HTML]{228B22}{$\uparrow$}16\\
        &   672	    &		\textcolor[HTML]{228B22}{$\uparrow$}28	&		\textcolor[HTML]{228B22}{$\uparrow$}13\\
        &   1344	&		\textcolor[HTML]{228B22}{$\uparrow$}9	&		\textcolor[HTML]{228B22}{$\uparrow$}5\\
        \midrule
        \multirow{3}{*}{\textit{CLIP}}
        &   336	    &		\textcolor[HTML]{228B22}{$\uparrow$}16	&		\textcolor[HTML]{228B22}{$\uparrow$}2\\
        &   672	    &		\textcolor[HTML]{228B22}{$\uparrow$}9	&		\textcolor[HTML]{228B22}{$\uparrow$}2\\
        &   1344	&		\textcolor[HTML]{228B22}{$\uparrow$}11	&	\textcolor{red}{$\downarrow$}1\\
        \bottomrule
    \end{tabularx}
    \caption{Difference in sccuracy between \textit{regular} (i.e., Table \ref{tab:coco-rq1-acc-full}) and \textit{small} objects (i.e., resized by $\frac{1}{4}$) when considering COCO objects.}
    \label{tab:rq1-delta-coco}
\end{table}

\begin{table*}[t]
    \centering
    \small
    \begin{tabularx}{0.66\textwidth}{lrcccc}
        \toprule
        \textbf{M\scriptsize{odel}} & \textbf{I\scriptsize{mage} \small{S}\scriptsize{ize}} & \textbf{S\scriptsize{hape}} & \textbf{C\scriptsize{olor}} & \textbf{S\scriptsize{heen}} & \textbf{P\scriptsize{osition}} \\
        \midrule
        \multirow{3}{*}{\textit{LLaVA-NeXT 7B}}  
        &   336	    &	$1 \pm 0.00$	&	$1 \pm 0.00$	&	$1 \pm 0.00$	&	$2 \pm 0.35$\\
        &   672	    &	$1 \pm 0.00$	&	$1 \pm 0.00$	&	$1 \pm 0.00$	&	$2 \pm 0.35$\\
        &   1344	&	$1 \pm 0.00$	&	$1 \pm 0.00$	&	$1 \pm 0.00$	&	$2 \pm 0.35$\\
        \midrule
        \multirow{3}{*}{\textit{LLaVA-NeXT 13B}}	            
        &   336	    &	$1 \pm 0.00$	&	$1 \pm 0.00$	&	$1 \pm 0.00$	&	$2 \pm 0.47$\\
        &   672	    &	$1 \pm 0.00$	&	$1 \pm 0.00$	&	$1 \pm 0.00$	&	$2 \pm 0.46$\\
        &   1344	&	$1 \pm 0.00$	&	$1 \pm 0.00$	&	$1 \pm 0.00$	&	$2 \pm 0.46$\\
        \midrule
        \multirow{3}{*}{\textit{Molmo-O 7B}}
        &   336	    &	$1 \pm 0.00$	&	$1 \pm 0.00$	&	$1 \pm 0.00$	&	$2 \pm 0.30$\\
        &   672	    &	$1 \pm 0.00$	&	$1 \pm 0.00$	&	$1 \pm 0.00$	&	$2 \pm 0.29$\\
        &   1344	&	$1 \pm 0.00$	&	$1 \pm 0.00$	&	$1 \pm 0.00$	&	$2 \pm 0.29$\\
        \midrule
        \multirow{3}{*}{\textit{Qwen2-VL 7B}}
        &   336	    &	$1 \pm 0.07$	&	$1 \pm 0.06$	&	$1 \pm 0.48$	&	$2 \pm 0.29$\\
        &   672	    &	$1 \pm 0.01$	&	$1 \pm 0.01$	&	$1 \pm 0.44$	&	$2 \pm 0.34$\\
        &   1344	&	$1 \pm 0.05$	&	$1 \pm 0.01$	&	$1 \pm 0.29$	&	$2 \pm 0.33$\\
        \bottomrule
    \end{tabularx}
    \caption{Answers length (i.e., average number of tokens with its standard deviation) of each model when considering different image sizes.}
    \label{tab:ans-len-instr}
\end{table*}

\begin{table*}[t]
    \small
    \centering
    \begin{tabularx}{0.55\textwidth}{lrcccc}
        \toprule
        \textbf{M\scriptsize{odel}} & \textbf{I\scriptsize{mage} \small{S}\scriptsize{ize}} & \textbf{S\scriptsize{hape}} & \textbf{C\scriptsize{olor}} & \textbf{S\scriptsize{heen}} & \textbf{P\scriptsize{osition}} \\
        \midrule
        \multirow{3}{*}{\textit{LLaVA-NeXT 7B}}  
        &   336 	&	0.00	&	0.69	&	0.00	&	0.01\\
        &   672 	&	0.00	&	0.35	&	0.00	&	0.00\\
        &   1344	&	0.00	&	0.33	&	0.00	&	0.00\\
        \midrule
        \multirow{3}{*}{\textit{LLaVA-NeXT 13B}}	            
        &   336  	&	0.00	&	6.76	&	0.00	&	0.00\\
        &   672  	&	0.00	&	2.04	&	0.00	&	0.00\\
        &   1344	&	0.00	&	1.84	&	0.00	&	0.00\\
        \midrule
        \multirow{3}{*}{\textit{Molmo-O 7B}}
        &   336 	&	0.00	&	0.00	&	0.00	&	0.00\\
        &   672 	&	0.00	&	0.00	&	0.00	&	0.00\\
        &   1344	&	0.00	&	0.00	&	0.00	&	0.00\\
        \midrule
        \multirow{3}{*}{\textit{Qwen2-VL 7B}}
        &   336 	&	0.70	&	0.98	&	1.57	&	0.30\\
        &   672 	&	0.30	&	0.89	&	1.63	&	0.33\\
        &   1344	&	0.27	&	1.27	&	1.39	&	0.36\\
        \bottomrule
    \end{tabularx}
    \caption{Percentage of "other" answers (i.e., answers outside the provided set of options, or multiple options from the set) generated by a model when considering different image sizes.}
    \label{tab:others}
\end{table*}

\begin{table*}[t]
    \small
    \centering
    \begin{tabularx}{0.57\textwidth}{lrrrrr}
        \toprule
        \textbf{M\scriptsize{odel}} & \textbf{I\scriptsize{mage} \small{S}\scriptsize{ize}} & \textbf{S\scriptsize{hape}} & \textbf{C\scriptsize{olor}} & \textbf{S\scriptsize{heen}} & \textbf{P\scriptsize{osition}} \\
        \midrule
        \textit{Random Baseline}	                    
        &		    &	25	&	17	&	50	&	11\\
        \midrule
        \multirow{3}{*}{\textit{LLaVA-NeXT 7B}}  
        &   336	    &	98	&	90	&	50	&	37\\
        &   672	    &	98	&	88	&	50	&	42\\
        &   1344	&	97	&	88	&	50	&	42\\
        \midrule
        \multirow{3}{*}{\textit{LLaVA-NeXT 13B}}	            
        &   336	    &	100	&	71	&	57	&	37\\
        &   672	    &	97	&	76	&	64	&	47\\
        &   1344	&	97	&	76	&	67	&	47\\
        \midrule
        \multirow{3}{*}{\textit{Molmo-O 7B}}
        &   336	    &	100	&	94	&	54	&	64\\
        &   672	    &	100	&	98	&	59	&	62\\
        &   1344	&	100	&	98	&	59	&	62\\
        \midrule
        \multirow{3}{*}{\textit{Qwen2-VL 7B}}
        &   336 	&	98	&	98	&	55	&	52\\
        &   672 	&	99	&	99	&	60	&	61\\
        &   1344	&	99	&	98	&	61	&	64\\
        \midrule
        \multirow{3}{*}{\textit{CLIP}}
        &   336 	&	99	&	89	&	50	&	13\\
        &   672 	&	95	&	95	&	49	&	14\\
        &   1344	&	95	&	95	&	49	&	14\\
        \bottomrule
    \end{tabularx}
    \caption{Accuracy (\%) of each model when considering visual scenes containing a single object and querying about its \textit{shape}, \textit{color}, \textit{sheen}, and \textit{absolute position} (w.r.t. the background).}
    \label{tab:rq1-acc-full}
\end{table*}

\begin{table}[t]
    \small
    \centering
    \begin{tabularx}{\columnwidth}{lrrrr}
        \toprule
        \multirow{2}{*}{\textbf{M\scriptsize{odel}}} & \multicolumn{4}{c}{\textbf{S\scriptsize{hape}}} \\
        \cmidrule{2-5}
        & \texttt{S\scriptsize{quare}} & \texttt{C\scriptsize{ircle}} & \texttt{T\scriptsize{riangle}} & \texttt{S\scriptsize{tar}} \\
        \midrule
        \textit{LLaVA-NeXT 7B}  
        &	96	&	96	&	100	&	100\\
        \midrule
        \textit{LLaVA-NeXT 13B}	            
        &	97	&	98	&	96	&	96\\
        \midrule
        \textit{Molmo-O 7B}
        &	100	&	100	&	100	&	100\\
        \midrule
        \textit{Qwen2-VL 7B}
        &	98	&	99	&	99	&	100\\
        \midrule
        \textit{CLIP}
        &	99	&	91	&	93	&   98\\
        \bottomrule
    \end{tabularx}
    \caption{F1-Score (\%) of each model when considering visual scenes containing a single object and querying about its \textit{shape}. Images have $672\times672$ pixels.}
    \label{tab:f1-shape}
\end{table}

\begin{table}[t!]
    \small
    \centering
    \begin{tabularx}{0.67\columnwidth}{lrrr}
        \toprule
        \multirow{2}{*}{\textbf{M\scriptsize{odel}}} & \multicolumn{2}{c}{\textbf{S\scriptsize{heen}}} \\
        \cmidrule{2-3}
        & \texttt{M\scriptsize{atte}} & \texttt{G\scriptsize{lossy}} \\
        \midrule
        \textit{LLaVA-NeXT 7B}  
        &	1	&	67\\
        \midrule
        \textit{LLaVA-NeXT 13B}	            
        &	66	&	61\\
        \midrule
        \textit{Molmo-O 7B}
        &	31	&	71\\
        \midrule
        \textit{Qwen2-VL 7B}
        &	44	&	70\\
        \midrule
        \textit{CLIP}
        &	66	&	0\\
        \bottomrule
    \end{tabularx}
    \caption{F1-Score (\%) of each model when considering visual scenes containing a single object and querying about its \textit{sheen}. Images have $672\times672$ pixels.}
    \label{tab:f1-sheen}
\end{table}

\begin{table*}[t!]
    \small
    \centering
    \begin{tabularx}{0.84\textwidth}{lrrrrrrrrrrr}
        \toprule
        \multirow{2}{*}{\textbf{M\scriptsize{odel}}} & \multicolumn{3}{c}{\textbf{T\scriptsize{op}}} & & \multicolumn{3}{c}{\textbf{C\scriptsize{enter}}} & & \multicolumn{3}{c}{\textbf{B\scriptsize{ottom}}} \\
        \cmidrule{2-4}
        \cmidrule{6-8}
        \cmidrule{10-12}
        & \texttt{L\scriptsize{eft}} & \texttt{C\scriptsize{enter}} & \texttt{R\scriptsize{ight}} & & \texttt{L\scriptsize{eft}} & \texttt{C\scriptsize{enter}} & \texttt{R\scriptsize{ight}} & & \texttt{L\scriptsize{eft}} & \texttt{C\scriptsize{enter}} & \texttt{R\scriptsize{ight}} \\
        \midrule
        \textit{LLaVA-NeXT 7B}  
        &	61	&	43	&	56	&&	6	&	30	&	4	&&	45	&	41	&	53\\
        \midrule
        \textit{LLaVA-NeXT 13B}	            
        &	58	&	44	&	59	&&	7	&	43	&	23	&&	62	&	42	&	62\\
        \midrule
        \textit{Molmo-O 7B}
        &	74	&	54	&	74	&&	40	&	53	&	46	&&	72	&	58	&	76\\
        \midrule
        \textit{Qwen2-VL 7B}
        &	70	&	64	&	75	&&	15	&	66	&	19	&&	70	&	59	&	67\\
        \midrule
        \textit{CLIP}
        &	23	&	14	&	5	&&	20	&	0	&	0	&&	1	&	4	&	12\\
        \bottomrule
    \end{tabularx}
    \caption{F1-Score (\%) of each model when considering visual scenes containing a single object and querying about its \textit{absolute position} (w.r.t. the background). Images have $672\times672$ pixels.}
    \label{tab:f1-abs}
\end{table*}

\begin{table*}[t!]
    \small
    \centering
    \begin{tabularx}{0.75\textwidth}{lrrrrrrrrrr}
        \toprule
        \multirow{2}{*}{\textbf{M\scriptsize{odel}}} & \multicolumn{2}{c}{\textbf{A\scriptsize{bove}}} & & \multicolumn{4}{c}{\textbf{D\scriptsize{irectly}}} & & \multicolumn{2}{c}{\textbf{B\scriptsize{elow}}} \\
        \cmidrule{2-3}
        \cmidrule{5-8}
        \cmidrule{10-11}
        & \texttt{L\scriptsize{eft}} & \texttt{R\scriptsize{ight}} & & \texttt{L\scriptsize{eft}} & \texttt{A\scriptsize{bove}} & \texttt{R\scriptsize{ight}} & \texttt{B\scriptsize{elow}} & & \texttt{L\scriptsize{eft}} & \texttt{R\scriptsize{ight}} \\
        \midrule
        \textit{LLaVA-NeXT 7B}
        &	34	&	43	&&	11	&	13	&	16	&	17	&&	18	&	13\\
        \midrule
        \textit{LLaVA-NeXT 13B}
        &	55	&	51	&&	10	&	2	&	16	&	12	&&	37	&	31\\
        \midrule
        \textit{Molmo-O 7B}
        &	15	&	35	&&	16	&	17	&	24	&	15	&&	27	&	33\\
        \midrule
        \textit{Qwen2-VL 7B}
        &	54	&	51	&&	20	&	15	&	21	&	18	&&	52	&	52\\
        \midrule
        \textit{CLIP}
        &	33	&	0	&&	0	&	0	&	0	&	0	&&	0	&	0\\
        \bottomrule
    \end{tabularx}
    \caption{F1-Score (\%) of each model when considering visual scenes containing two objects and querying about the \textit{relative position} of one object w.r.t. another. Images have $672\times672$ pixels.}
    \label{tab:f1-rel-pos}
\end{table*}

\begin{table}[t!]
    \small
    \centering
    \begin{tabularx}{0.93\columnwidth}{lrrr}
        \toprule
        \textbf{M\scriptsize{odel}} & \textbf{I\scriptsize{mage} \small{S}\scriptsize{ize}} & \textbf{C\scriptsize{ategory}} & \textbf{P\scriptsize{osition}} \\
        \midrule
        \textit{Random Baseline}
        &		    &	33	&	11\\
        \midrule
        \multirow{3}{*}{\textit{LLaVA-NeXT 7B}}
        &   336 	&	55	&	23\\
        &   672 	&	91	&	35\\
        &   1344	&	91	&	37\\
        \midrule
        \multirow{3}{*}{\textit{LLaVA-NeXT 13B}}
        &   336 	&	56	&	37\\
        &   672 	&	90	&	45\\
        &   1344	&	80	&	51\\
        \midrule
        \multirow{3}{*}{\textit{Molmo-O 7B}}
        &   336 	&	34	&	45\\
        &   672 	&	69	&	59\\
        &   1344	&	70	&	53\\
        \midrule
        \multirow{3}{*}{\textit{Qwen2-VL 7B}}
        &   336 	&	60	&	30\\
        &   672 	&	86	&	49\\
        &   1344	&	97	&	52\\
        \midrule
        \multirow{3}{*}{\textit{CLIP}}
        &   336 	&	65	&	13\\
        &   672 	&	67	&	16\\
        &   1344	&	67	&	15\\
        \bottomrule
    \end{tabularx}
    \caption{Accuracy (\%) of each model when considering visual scenes containing different objects from COCO (i.e., zebra, giraffe, and elephant) and querying about their \textit{category} and \textit{absolute position} (w.r.t. the background).}
    \label{tab:coco-rq1-acc-full}
\end{table}

\begin{table}[t!]
    \small
    \centering
    \begin{tabularx}{0.88\columnwidth}{lrrr}
        \toprule
        \multirow{2}{*}{\textbf{Model}} & \multicolumn{3}{c}{\textbf{Category}} \\
        \cmidrule{2-4}
        & \texttt{G\scriptsize{iraffe}} & \texttt{E\scriptsize{lephant}} & \texttt{Z\scriptsize{ebra}} \\
        \midrule
        \textit{LLaVA-NeXT 7B}  
        &	94	&	84	&	92\\
        \midrule
        \textit{LLaVA-NeXT 13B}	            
        &	88	&	59	&	86\\
        \midrule
        \textit{Molmo-O 7B}
        &	99	&	16	&	69\\
        \midrule
        \textit{Qwen2-VL 7B}
        &	96	&	95	&	99\\
        \midrule
        \textit{CLIP}
        &	70	&	0	&	94\\
        \bottomrule
    \end{tabularx}
    \caption{F1-Score (\%) of each model when considering visual scenes containing different objects from COCO (i.e., zebra, giraffe, and elephant) and querying about their \textit{category}. Images have $1344\times1344$ pixels.}
    \label{tab:f1-coco-cat}
\end{table}

\end{document}